# AN EXHAUSTIVE STUDY OF THE WORKSPACES TOPOLOGIES OF ALL 3R ORTHOGONAL MANIPULATORS WITH GEOMETRIC SIMPLIFICATIONS

Mazen Zein, Philippe Wenger and Damien Chablat

ABSTRACT- This paper analyses the workspace of the three-revolute orthogonal manipulators that have at least one of their DH parameters equal to zero. These manipulators are classified into different groups with similar kinematic properties. The classification criteria are based on the topology of the workspace. Each group is evaluated according to interesting kinematic properties such as the size of the workspace subregion reachable with four inverse kinematic solutions, the existence and the size of voids, and the size of the regions of feasible paths in the workspace.

## I. INTRODUCTION

Today, most serial industrial manipulators are of the PUMA type, they have a vertical revolute joint followed by two parallel joints and a spherical wrist. Another category of serial manipulators exists, these manipulators are called the orthogonal manipulators, they have three orthogonal joint axes. The IRB 6400C launched by the ABB-Robotics pertains to this category.

Unlike usual industrial manipulators, orthogonal manipulators may have many different kinematic features, such as cuspidality. A cuspidal manipulator is one that can change posture without meeting a singularity [1,2]. [3,4] provides conditions for a manipulator to be noncuspidal. A general necessary and sufficient condition for a 3-DOF manipulator to be cuspidal was established in [5], which is the existence of at least one point, called cusp point, in the workspace where the inverse kinematics admits three equal solutions.

Orthogonal manipulators may have different global kinematic properties according to their links and joint offsets lengths, so it is is interesting to classify them, given that such manipulators can be binary or quaternary, generic or non-generic, cuspidal or noncuspidal. [6] established a categorisation of all generic 3R manipulators based on homotopy classes. More recently, [7] attempted the classification of 3R orthogonal manipulators with no offset on their last joint. Three surfaces were found to divide the manipulator parameters space into cells with constant number of cusp points. The equations of these surfaces were derived as polynomials in the DH-parameters using Groebner Bases. The work of [7] was completed in [8] to take into account additional features in the classification like genericity and the number of aspects. [9] established a classification of 3R orthogonal manipulators with no offset on their last joint, based on the work of [8], according to the number of cusps and node points, the parameters space was divided in 9 cells where the manipulators have the same number of cusps and nodes in their workspace.

[9] classified only the family of 3R orthogonal manipulators with no offset on their last joint. About ten remaining families of 3R manipulators with at least one of its DH parameters equal to zero have not been classified yet. And, since the majority of the industrial manipulators have at least one parameter equal to zero, it is very important to classify them.

The main purpose of this paper is to analyse the workspace of the three-revolute orthogonal manipulators that have at least one of their DH parameters equal to zero. These manipulators are classified into different groups with similar kinematic properties. The classification criteria are based on the topology of the workspace. Each group is evaluated according to interesting kinematic properties such as the size of the workspace subregion reachable with four inverse kinematic solutions, the existence and the size of voids, and the size of the regions of feasible paths.

Next section of this article presents the familties of manipulators under study and recalls some preliminary results. The classifications are established in section III. In section IV a ranking of the groups found is established and section V concludes this paper.

## II. PRELIMINARIES

### II.1 Orthogonal manipulators

The orthogonal manipulators studied have three orthogonal revolute joint axes, their DH-parameters are $d_2$, $d_3$, $d_4$, $r_2$ and $r_3$, the angle parameters $\alpha_2$ and $\alpha_3$ are set to –90° and 90°, respectively. The three joint variables are referred to as $\theta_1$, $\theta_2$ and $\theta_3$, respectively, they will be assumed unlimited in this study. The position of the end-tip (or wrist center) is defined by the three Cartesian coordinates $x$, $y$ and $z$ of the operation point P with respect to a reference frame (O, **X**,**Y**, **Z**) attached to the manipulator base. Figure 1 shows the architecture of the manipulators under study in the zero configuration ($\theta_1=\theta_2=\theta_3=0$).

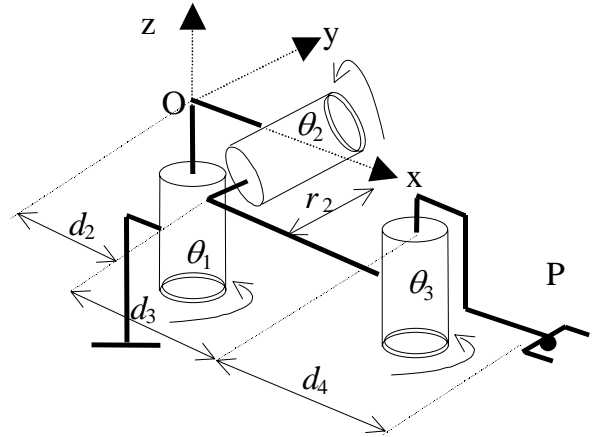

*Fig 1. 3R Orthogonal manipulator*

### II.2 Problem Formulation

The topology and the geometry of the singularities of a manipulator are an important way to obtain its global kinematic properties. Thus, the classification criteria will be the topology of the singularity curves in a cross section of the workspace, which is defined by the number of cusp points and the number of node points that appear on these curves.

A cusp point is a point of the workspace of a manipulator that has three coincident inverse kinematic solutions [5]. A 3-DOF positioning manipulator can change its posture without meeting a singularity only by encountering a cusp point [10].

A node point is a point of the workspace of a manipulator that has two pairs of coincident inverse kinematic solutions [8]. Two branch of singularities intersect at a node point.

The existence of cusps and nodes can be determined from the direct kinematic equations, which can be written as a polynomial in $t = \tan\left(\theta_3 / 2\right)$:

$$P(t) = at^4 + bt^3 + ct^2 + dt + e \tag{1}$$

where $a$, $b$, $c$, $d$, $e$ are functions of the design parameters $d_2$, $d_3$, $d_4$, $r_2$ and $r_3$ and the variables $x$, $y$ and $z$.

The manipulator admits a node (resp. a cusp) in its workspace if and only if $P(t)$ admits two pairs of equal roots (resp. real triple roots) [8].

### II.3 Particular manipulators

The manipulators studied and classified in [8] and [9] only considered the case $r_3$ =0. Here, we will treat all the possible combinations of manipulator with at least one DH parameter equal to zero. The different combinations are depicted by the following tree, which yields ten cases to consider (Figure 2). Note that $d_4$ cannot be equal to zero since the resulting manipulator would be always singular.

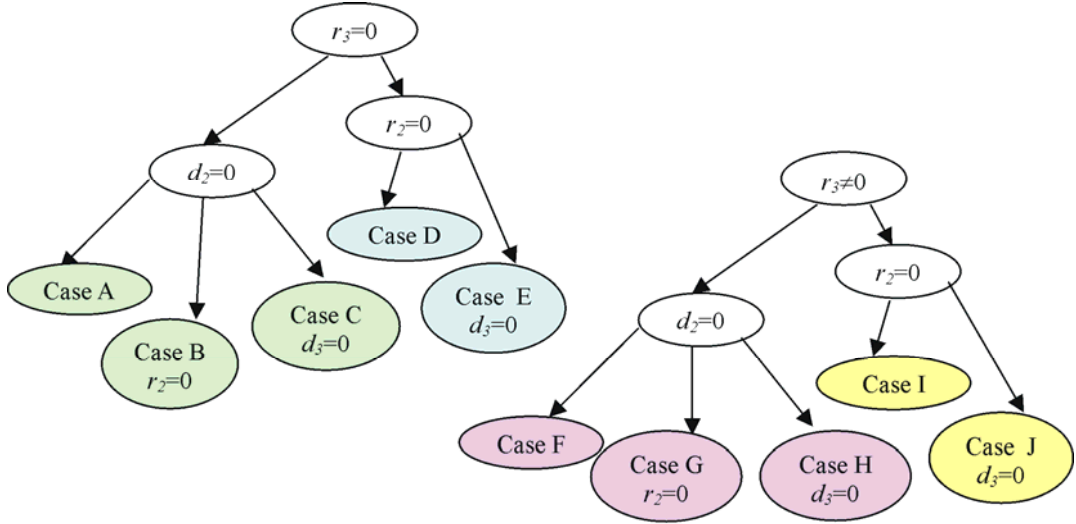

*Fig 2. Possible combinations of manipulators parameters with at least one parameter equal to zero*

## III. CLASSIFICATIONS OF EACH PATICULAR CASE

In this section, we classify each case shown in Fig. 2, to obtain groups of manipulators having similar kinematic properties. We know from [11] that when $d_2=0$ or $r_2=0$, the inverse kinematic polynomial can be written as a quadratics. Thus these manipulators cannot have cusp points in their workspaces, because, their inverse kinematic polynomial cannot admit triple roots. As a consequence, the classification criteria will be only the number of node points in the workspace. The parameters space of each case will be divided by different curves, called transition curves, into a number of zones. The equations of the transition curves will be computed from those found in [9] and [12], wich are:

$$(E1): d_4 = \frac{1}{2}(a - b)$$

$$(E2): d_4 = d_3 \qquad \text{for manipulators with } r_3=0 \text{ [9].}$$

$$(E3): d_4 = \frac{1}{2}(a + b)$$

$$(E1): d_4 = \sqrt{\frac{1}{2}\left[1 + r_2^2 + d_3^2\left(1 - \left(\frac{d_3}{r_2} + \frac{\ddot{o}}{\emptyset}\right) + \left(\frac{d_3}{r_2} - r_3^2 - ab\left|\frac{d_3}{r_2} - 1\right|\right)\right]\right]}$$

$$(E2): d_4 = \sqrt{\frac{1}{2}\left[1 + r_2^2 + d_3^2\left(1 - \left(\frac{d_3}{r_2} + \frac{\ddot{o}}{\emptyset}\right) + \left(\frac{d_3}{r_2} - r_3^2 + ab\left|\frac{d_3}{r_2} - 1\right|\right)\right]\right]}$$

for manipulators with $r_3 \neq 0$ [12]

with $a = \sqrt{(d_3 + d_2)^2 + r_2^2}$ et $b = \sqrt{(d_3 - d_2)^2 + r_2^2}$

Below, we expose the classification of each case, each zone of the parameters space corresponds to a group of manipulators having the same number of nodes. For each group the workspace is characterized in a half-cross section by the singular curves and the workspace topology is defined by the number of nodes that appear on these singular curves.

### III.1 Case A ($d_2=0$, $r_2 \neq 0$, $d_3 \neq 0$ et $r_3=0$)

These manipulators are such that $d_2=0$ and $r_3=0$. In this case, (E1) does not exist because $\alpha = \beta$ when $d_2=0$, which is not possible since $d_4 \neq 0$. Thus, the parameters space is divided, by the transition curves (E2) and (E3), into three zones ZONE 1, ZONE 2 and ZONE 3 collecting the manipulators with 0, 2 and 4 node points in their workspaces respectively. The equations of the transition curves are derived geometrically and from the equations found by [9], they are as follows:

(E2): $d_4 = d_3$

(E3): $d_4 = \sqrt{d_3^2 + r_2^2}$

Fig. 3 shows, for $r_2=1$, the parameters space with three zones, the transition curves (E1) and (E2), the workspace of a manipulator of each zone, and for each workspace the regions with four inverse kinematic solutions IKS are colored with dark gray, those with 2 IKS are in light gray, and the singularity curves in blue. <span style="color:red">ATTENTION CHANGER SUR LA FIGURE E1 ET E2</span>

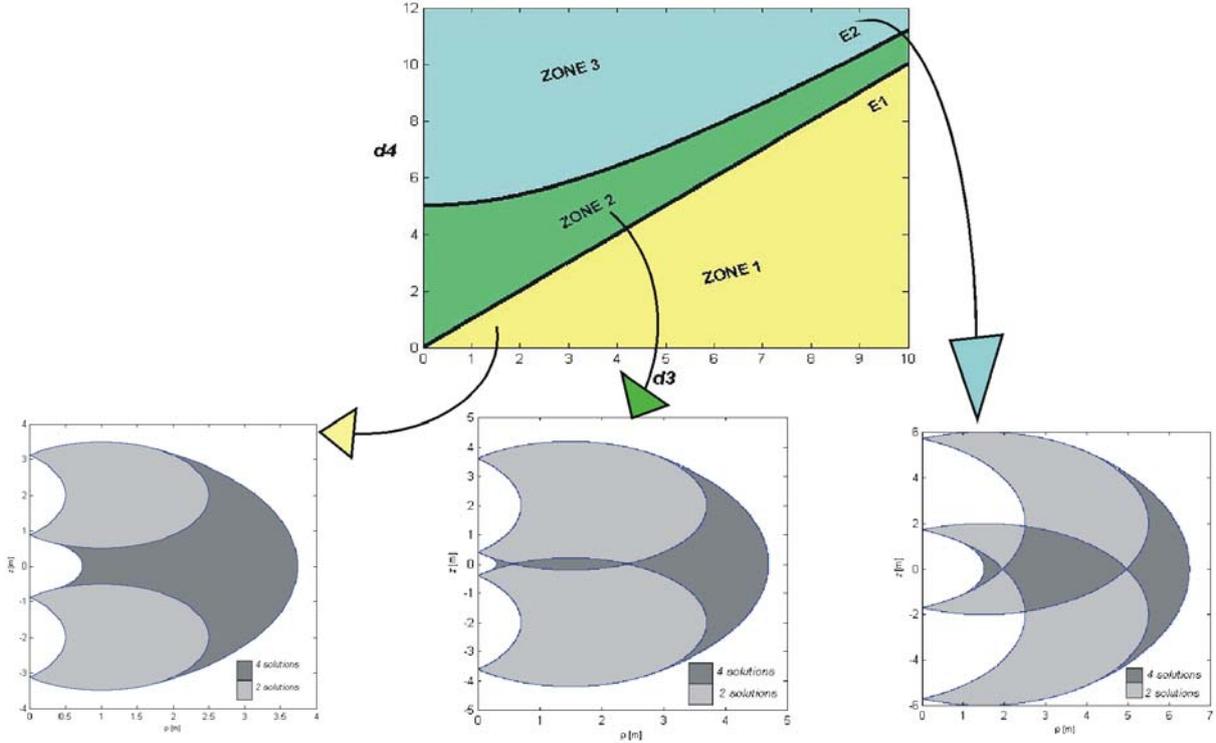

*Fig 3. Parameters space and workspaces of the manipulators having the following parameters from left to right: $d_3=2$, $d_4=1.5$ and $r_2=1$; $d_3=2$, $d_4=2.2$ and $r_2=1.5$; $d_3=2$, $d_4=3$ and $r_2=1$*

In this case, we have three groups of manipulators, which we call A1, A2 and A3 corresponding to ZONE 1, ZONE 2 and ZONE 3 respectively.

- **Group A1** (Fig. 3, left): The manipulators of this group have $d_3>d_4$. Their workspace admits no voids nor node points, they are composed of three regions, two of them are reachable with 2 IKS and the other one with 4 IKS. Any path is feasible throughout the workspace.

- **Group A2** (Fig. 3, middle): The manipulators of this group have $d_3<d_4<\sqrt{d_3^2 + r_2^2}$. Their workspace admits 2 node points and no voids. Any path is feasible throughout the workspace.

- **Group A3** (Fig. 3, right): The manipulators of this group have $d_4 > \sqrt{d_3^2 + r_2^2}$. Their workspace admits 4 node points and no voids. Any path is feasible throughout a significant region of the workspace.

### III.2 Case B ($d_2=0$, $r_2=0$, $d_3\neq0$ et $r_3=0$)

The manipulators of this case are such that $d_2=0$, $r_2=0$ and $r_3=0$. In this case, (E3) does not exist and (E2) is equivalent to (E1). Thus the parameters space is divided, by the transition curve (E1), into two zones ZONE 1 and ZONE 2 collecting the manipulators with 0 and 1 node points in their workspaces respectively. The equation of the transition curve is derived geometrically and from the equations found by [9]:

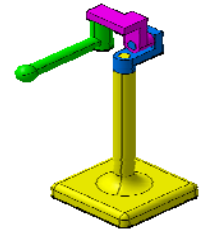

(E1): $d_3 = d_4$

The figure below shows the parameters space with two zones, the transition curve (E1), the workspace of a manipulator of each zone.

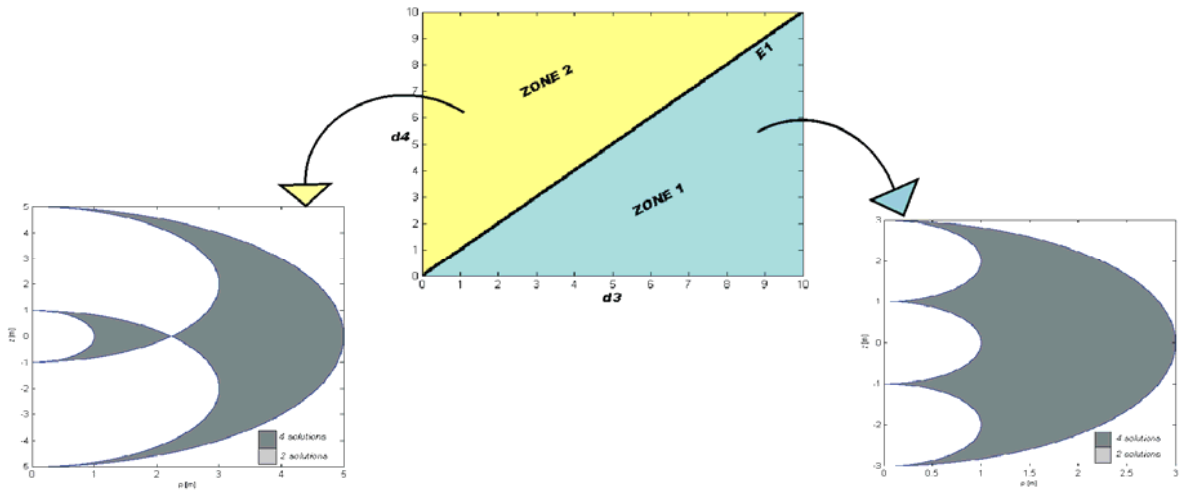

*Fig 4. Parameters space and workspaces of the manipulators having the following parameters from left to right: $d_3=2$ and $d_4=3$; $d_3=2$ and $d_4=1$.*

In this case, we have two groups of manipulators, which we call B1 and B2 corresponding to ZONE 1 and ZONE 2 respectiveley.

- **Group B1** (Fig. 4, right)**:** The manipulators of this group have $d_3>d_4$. Their workspace does not admit any voids and node points; they are composed of only one region, which is reachable with 4 IKS. Any path is feasible throughout the workspace.

- **Group B2** (Fig. 4, left)**:** The manipulators of this group have $d_4 > d_3$. Their workspace admits 1 node point and no voids. They are composed of two separated regions of feasible paths. Large holes appear in the workspaces of these manipulators.

### III.3 Case C ($d_2=0$, $r_2\neq0$, $d_3=0$ et $r_3=0$)

The manipulators of this case are such that $d_2=0$, $d_3=0$ and $r_3=0$. They do not have any node point in their workspace. Figure 5 shows the workspace of such a manipulator. In this case, we have only one group of manipulators, which we call Group C.<span style="color:red">PB: ON DEVRAIT TROUVER d4=r2 QUAND ON REMPLACE</span>

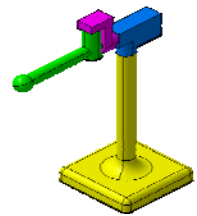

- **Group C** (Fig. 5)**:** The manipulators of this group do not admit voids nor node points in their workspace, which are composed of only one region reachable with 4 IKS. Any path is feasible throughout the workspace. This group of manipulators is very interesting.

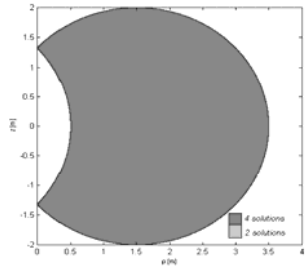

*Fig 5. Workspace of the manipulator having the following parameters $d_4=2$ and $r_2=1.5$.*

### III.4 Case D ($d_2 \neq 0$, $r_2=0$, $d_3 \neq 0$ et $r_3=0$)

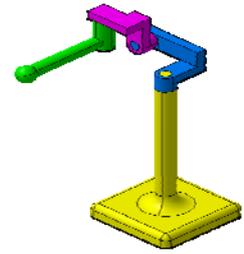

The manipulators of this case are such that $r_2=0$ and $r_3=0$. The parameters space is divided, by the transition curves (E1), (E2) and (E3) into five zones ZONE 1, ZONE 2, ZONE 3, ZONE 4 and ZONE 5 collecting the manipulators with 2, 0, 1, 2 and 0 nodes points in their workspaces respectively. The equations of the transition curves derived geometrically and from the equations found by [9] are as follows:<span style="color:red">PROBLEME DANS CES FORMULES ON N'OBTIENT PAS CA QUAND ON REMPLACE</span>

(E1): $d_4 = d_2$

(E2): $d_4 = d_3$ if $d_3$ and $d_4 > d_2$

(E3): $d_3 = d_2 = 1$

Figure 6 shows the parameters space with five zones, the transition curves (E1), (E2) and (E3), the workspace of a manipulator of each zone.

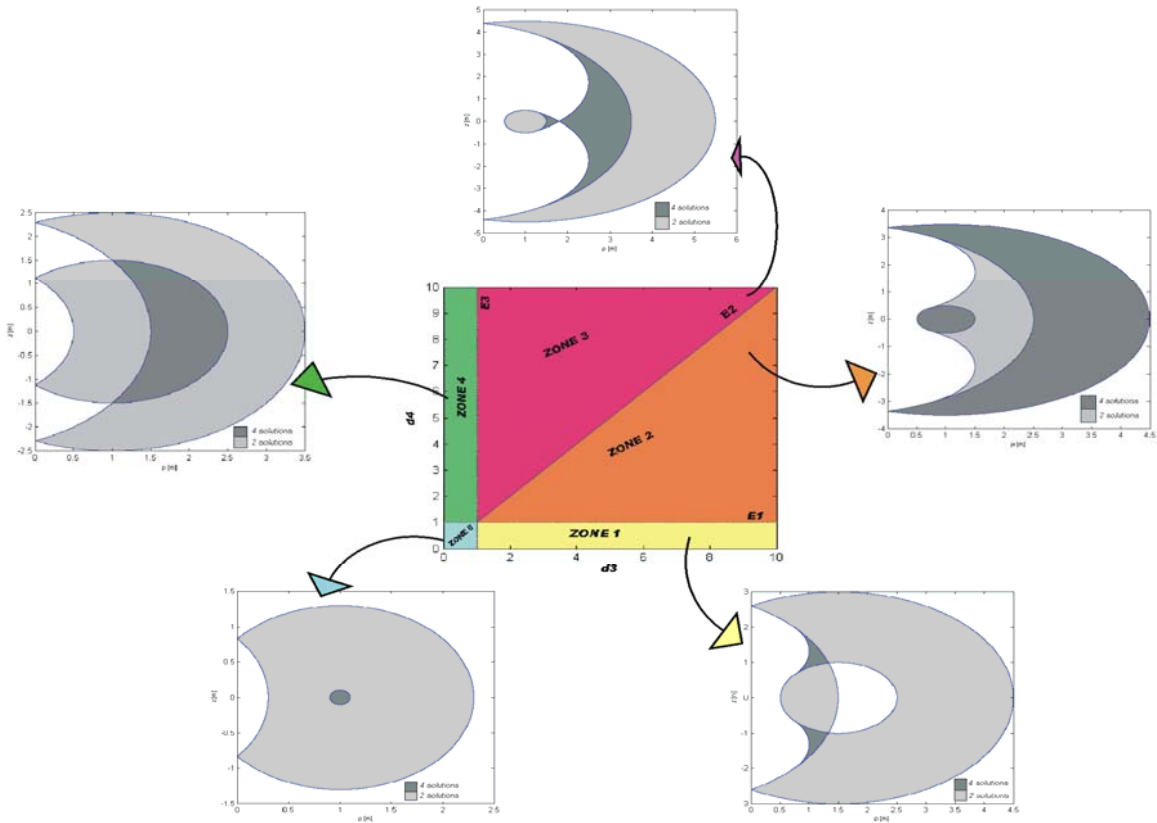

*Fig 6. Parameters space and workspaces of the manipulators having the following parameters: $d_3=1.4$, $d_4=0.7$ and $d_2=1$ (ZONE 1); $d_3=2$, $d_4=1.5$ and $d_2=1$ (ZONE 2); $d_3=2$, $d_4=2.5$ et $d_2=1$ (ZONE 3); $d_3=0.5$, $d_4=2$ and $d_2=1$(ZONE 4); $d_3=0.6$, $d_4=0.7$ et $d_2=1$(ZONE 5).*

In this case, we have five groups of manipulators, refered to as D1, D2, D3, D4 and D5 corresponding to ZONE 1, ZONE 2, ZONE 3, ZONE 4 and ZONE 5 respectiveley.

- **Group D1:** The manipulators of this group have $d_4 < d_2 < d_3$. Their workspace admits one void and 2 node points. The main part of their workspace is a region reachable 2 IKS. No region of feasible paths exists that covers all the workspace.

- **Group D2:** The manipulators of this group have $d_2 < d_4 < d_3$. Their workspace does not admit voids nor node points. The main part of their workspace is a region reachable 4 IKS.

- **Group D3:** The manipulators of this group have $d_2 < d_3 < d_4$. Their workspace admits 1 node point and no voids. No region of feasible paths exists that covers all the workspace. Large holes appear in the workspaces of these manipulators.

- **Group D4:** The manipulators of this group have $d_3 < d_2 < d_4$. Their workspace admits 2 node points and no voids. No region of feasible paths exists that covers all the workspace.

- **Group D5:** The manipulators of this group have $d_3, d_4 < d_2$. Their workspace does not admit voids nor node points. The main part of their workspace is a region reachable 2 IKS. Any path is feasible throughout the workspace.

### III.5 Case E ($d_2 \neq 0$, $r_2 = 0$, $d_3 = 0$ et $r_3 = 0$)

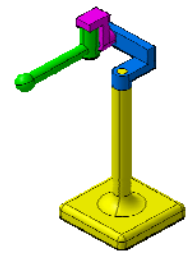

The manipulators of this case are such that $d_3 = 0$, $r_2 = 0$ and $r_3 = 0$. They do not have any node point in their workspace. The figure below shows the workspace of a manipulator. In this case, we have only one group of manipulators which we call Group E. <span style="color:red">PB: ON DEVRAIT TROUVER d4=d2 QUAND ON REMPLACE</span>

- **Group E** (Fig. 7)**:** The manipulators of this group are generic and do not admit voids nor node points in their workspace, which are composed of only one region reachable with 4 IKS. Any path is feasible throughout the workspace. This group is very interesting.

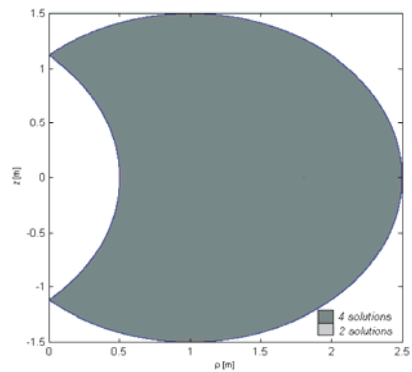

*Fig 7. Workspace of the manipulator having the following parameters $d_4 = 1.5$ and $d_2 = 1$*

### III.6 Case F ($d_2 = 0$, $r_2 \neq 0$, $d_3 \neq 0$ et $r_3 \neq 0$)

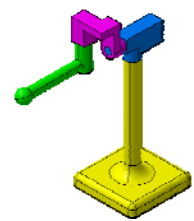

The manipulators of this case are such that $d_2 = 0$. They can have 0 or 2 node points in their workspace, according to the relation between their design parameters. As a consequence the parameters space is divided, by the transition curve (E1), into two zones ZONE 1 and ZONE 2 collecting the manipulators with 0 and 2 node points in their workspace respectively. The equation of the transition curve is derived geometrically and from the equations found by [12]:

(E1): $d_4 = \sqrt{d_3^2 + r_2^2}$

The figure below shows the parameters space with three zones, the transition curves (E1) and (E2), the workspace of a manipulator of each zone, and for each workspace the regions with four inverse kinematic solutions IKS are filled in dark gray, those with 2 IKS are in light gray.

In this case, there exists two groups of manipulators, refered to as F1 and F2 corresponding to ZONE 1 and ZONE 2, respectively.

- **Group F1** (Fig. 8, right)**:** The manipulators of this group have $d_4 < \sqrt{d_3^2 + r_2^2}$ . Their workspace does not admit voids nor node points, they are composed of three regions, two of them being reachable with 2 IKS and the other one with 4 IKS. Any path is feasible throughout the workspace.

- **Group F2** (Fig. 8, left)**:** The manipulators of this group have $d_4 > \sqrt{d_3^2 + r_2^2}$ . Their workspace admits 2 node points and no voids. Any path is feasible throughout a significant region of the workspace.

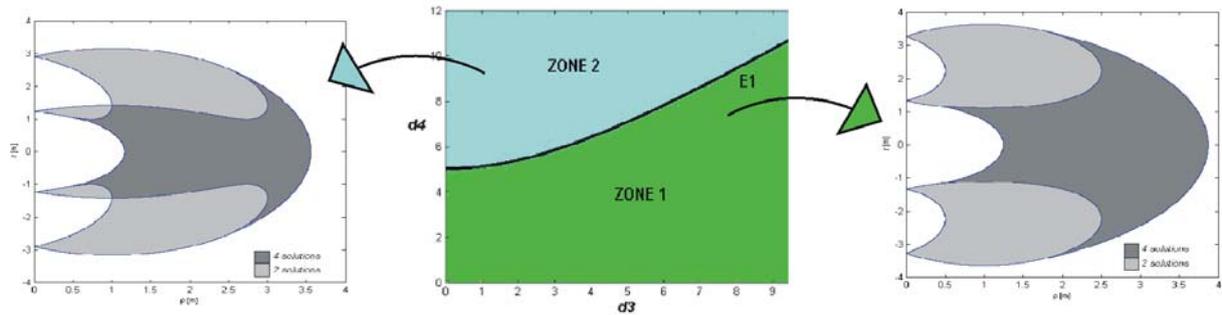

*Fig 8. Parameters space and workspaces of the manipulators having the following parameters from left to right: $d_3=2$, $d_4=1.5$, $r_2=1$ and $r_3=1$; $d_3=1$, $d_4=2$, $r_2=1$ and $r_3=1$.*

### III.7 Case G ($d_2=0$, $r_2=0$, $d_3\neq0$ et $r_3\neq0$)

The manipulators of this case are such that $d_2=0$ and $d_3=0$. They do not have any node point in their workspace. <span style="color:red">POURQUOI?</span> The figure below shows the workspace of a manipulator. In this case, we have only one group of manipulators which we call Group G.

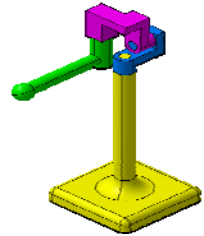

- **Group G** (Fig. 9)**:** The manipulators of this group do not admit voids nor node points in their workspace, which are composed of only one region reachable with 4 IKS. Any path is feasible throughout the workspace. Large holes appear in the workspace of these manipulators.

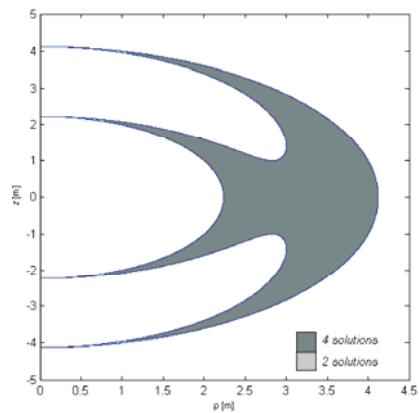

*Fig 9. Workspace of the manipulator having the following parameters $d_3=1$, $d_4=3$, $r_2=0$ and $r_3=1$*

### III.8 Case H ($d_2$=0, $r_2$≠0, $d_3$=0 et $r_3$≠0)

The manipulators of this case are such that $d_2$=0 and $d_3$=0. They do not have any node point nor voids in their workspace.POURQUOI?The figure below shows the workspace of a manipulator. In this case, we have only one group of manipulators which we call Group H.

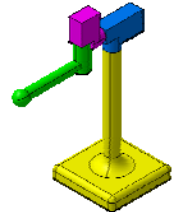

* **Group H** (Fig. 10)**:** The manipulators of this group do not admit voids nor node points in their workspace, which are composed of only one region reachable with 4 IKS Any path is feasible throughout the workspace. These manipulators have very large holes inside their workspace.

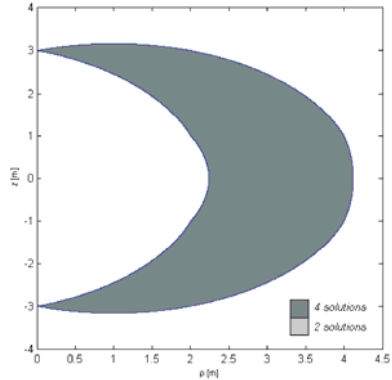

*Fig 10. Workspace of the manipulator having the following parameters $d_3$=0, $d_4$=1, $r_2$=3 and $r_3$=1*

### III.9 Case I ($d_2$≠0, $r_2$=0, $d_3$≠0 et $r_3$≠0)

The manipulators of this case are such that $r_2$=0. The parameters space is divided, by the transition curves (E1) and (E2) into four zones ZONE 1, ZONE 2, ZONE 3 and ZONE 4 collecting the manipulators with 0, 2, 0 and 2 nodes points in their workspaces respectively. The equations of the transition curves derived geometrically and from the equations found by [9] are as follows:

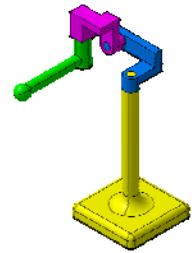

(E1): $d_4 = \delta$ with $\delta = \sqrt{1 + \dfrac{r_3^2}{d_3^2 - 1}}$

(E2): $d_3 = d_2$

Figure 11 shows the parameters space with $r_3$=0.5 with four zones, the transition curves (E1) and (E2), the workspace of a manipulator of each zone.
In this case, we have four groups of manipulators, which we call I1, I2, I3 and I4 corresponding to ZONE 1, ZONE 2, ZONE 3 and ZONE 4 respectively.

* **Group I1:** The manipulators of this group have $d_3 > d_2$ and $d_4 > \delta$. Their workspace does not admit voids nor node points. The main part of their workspace is a region reachable with 2 IKS. No region of feasible paths exists that covers all the workspace.

* **Group I2:** The manipulators of this group have $d_3 > d_2$ and $d_4 < \delta$. Their workspace admits 2 node points and one void. No region of feasible paths exists that covers all the workspace.

* **Group I3:** The manipulators of this group have $d_3 < d_2$ and $d_4 > \delta$. Their workspace admits one void and no node points. The main part of their workspace is a region reachable 2 IKS. Any path is feasible throughout the workspace.

* **Group I4:** The manipulators of this group have $d_3 < d_2$ and $d_4 < \delta$. Their workspace admits 2 node points and one void in their workspace. No region of feasible paths exists that covers all the workspace.

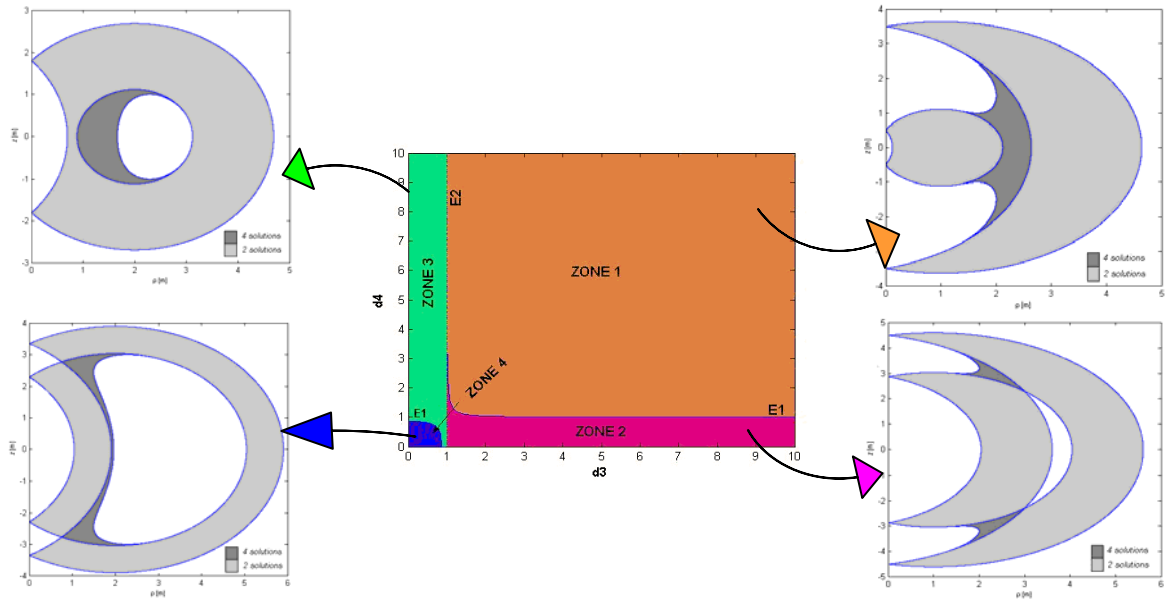

*Fig 11. Parameters space with $r_3$=0.5 and workspaces of the manipulators having the following parameters:*
*$d_2$=1, $d_3$=2.5 and $d_4$=1.5 (ZONE 1); $d_2$=1, $d_3$=3 and $d_4$=0.7 (ZONE 2);*
*$d_2$=1, $d_3$=0.5 and $d_4$=0.7 (ZONE 3); $d_2$=1, $d_3$=0.3 and $d_4$=2(ZONE 4).*

### III.10 Case J ($d_2 \neq 0$, $r_2$=0, $d_3$=0 et $r_3 \neq 0$)

The manipulators of this case have $r_2$=0 and $d_3$=0. They have one void and no node points in their workspace.<span style="color:red">POURQUOI?</span>The figure below shows the workspace of a manipulator. In this case, we have only one group of manipulators which we call Group J.

- **Group J** (Fig. 13)**:** The manipulators of this group admit one void and no node points in their workspace, which are composed of only one region reachable with 4 IKS. Any path is feasible throughout the workspace.

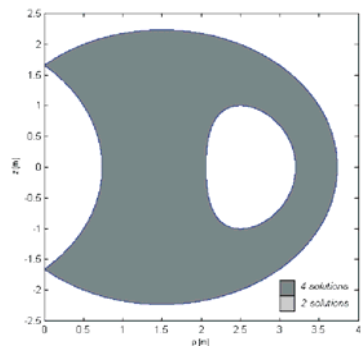

*Fig 12. Workspace of the manipulator having the following parameters $d_2$=1, $d_4$=2, and $r_3$=1*

### IV. CATEGORISATION OF THE GROUPS

In the previous section, we have classifed the ten cases of null parameters combinations. Twenty one different groups of manipulators have been found, each one contains manipulators having the same workspace topology and with similar global kinematics properties. Some groups have interesting properties, other ones have defective properties such as large holes and voids ins their workspace.

In this section, the groups in three classes will be ranked. The first one is the most interesting one, and the last one is the least interesting one. First, we expose all the groups and their properties in the table below, and after that it will be easier to rank them.

| Groups | Void | Node points | 4 IKS zone | Holes | Feasible paths zone |
|--------|------|-------------|------------|-------|---------------------|
| A1 | 0 | 0 | Intermediate | Small | All the workspace |
| A2 | 0 | 2 | Small | Small | All the workspace |
| A3 | 0 | 4 | Small | Intermediate | All the workspace |
| B1 | 0 | 0 | All the workspace | Intermediate | All the workspace |
| B2 | 0 | 1 | All the workspace | Big | All the workspace |
| C | 0 | 0 | All the workspace | Small | All the workspace |
| D1 | 1 | 2 | Small | Small | Big |
| D2 | 0 | 0 | Big | Small | Intermediate |
| D3 | 0 | 1 | Small | Intermediate | Intermediate |
| D4 | 0 | 2 | Small | Small | Big |
| D5 | 0 | 0 | Small | Small | All the workspace |
| E | 0 | 0 | All the workspace | Small | All the workspace |
| F1 | 0 | 0 | Intermediate | Small | All the workspace |
| F2 | 0 | 2 | Intermediate | Small | Big |
| G | 0 | 0 | All the workspace | Big | All the workspace |
| H | 0 | 0 | All the workspace | Intermediate | All the workspace |
| I1 | 0 | 0 | Small | Intermediate | Big |
| I2 | 1 | 2 | Small | Intermediate | Intermediate |
| I3 | 1 | 0 | Small | Small | All the workspace |
| I4 | 1 | 2 | Small | Small | All the workspace |
| J | 1 | 0 | All the workspace | Small | All the workspace |

*Tab. All groups and their kinematic properties*

From this table and from the workspace topologies of the manpulators shown before, we propose the following ranking:

1. ***First class manipulators***: Groups C and E.

2. ***Second class manipulators***: Groups A1, A2, A3, B1, D2, D3, D4, D5, F1, F2, H, I1 and J.

3. ***Third class manipulators***: Groups B2, D1, G, I2, I3 and I4.

The workspace of the manipulator of the first class is composed of one zone with 4 IKS, in which all paths are feasible. These manipulators do not have node nor internal singularity curves nor voids and cavities inside their workspaces. So they can achieve any motion without crossing a singularity, and any path without changing posture.

In conclusion, the manipulators having the length $d_3$ of the third link null, are the most interesting. By optimizing the parameters of the manipulators, we can always find interesting geometries even if their groups are not included in the first or second class.

V. CONCLUSIONS

In this article, an exhaustive study of the workspace topologies of all the families of 3R orthogonal manipulators with at least one null DH parameter is established and all these particular families were classified. Twenty one different groups of manipulators, having similar global kinematic properties and same workspace topologies, were found. A table showing the manipulators properties for each group was set up, and from this table a ranking of the groups, according to their interesting properties, was deduced.